# Live Target Detection with Deep Learning Neural Network and Unmanned Aerial Vehicle on Android Mobile Device

Ali Canberk ANAR, Erkan BOSTANCI, Mehmet Serdar GUZEL
SAAT Laboratory
Department of Computer Engineering
Ankara University
Ankara, Turkey
canberkanar@gmail.com, ebostanci@ankara.edu.tr, mguzel@ankara.edu.tr

*Abstract*—This paper describes the stages faced during the development of an Android program which obtains and decodes live images from DJI Phantom 3 Professional Drone and implements certain features of the TensorFlow Android Camera Demo application. Test runs were made and outputs of the application were noted. A lake was classified as seashore, breakwater and pier with the accuracies of 24.44%, 21.16% and 12.96% respectfully. The joystick of the UAV controller and laptop keyboard was classified with the accuracies of 19.10% and 13.96% respectfully. The laptop monitor was classified as screen, monitor and television with the accuracies of 18.77%, 14.76% and 14.00% respectfully. The computer used during the development of this study was classified as notebook and laptop with the accuracies of 20.04% and 11.68% respectfully. A tractor parked at a parking lot was classified with the accuracy of 12.88%. A group of cars in the same parking lot were classified as sports car, racer and convertible with the accuracies of 31.75%, 18.64% and 13.45% respectfully at an inference time of 851ms.

## I. INTRODUCTION

Unmanned aerial vehicle (UAV) is a type of aircraft which is remotely controlled via a control unit. UAVs are currently being used in many different areas such as border patrol, noticing terrorists, destruction of targets, intelligence, fire extinguishing, traffic control, filming, cargo services, agricultural activities, prevention of smuggling, and personal use.

The main aim of this study was to create a simple system which autonomously detects certain objects and informs the user about the situation. The idea was mainly inspired from the successful operations, a military grade UAV has accomplished on national defense.

In order to get a thought on the methods that will be developed to achieve our goals, literature research regarding to pre-done studies about the topic was done. The projects about Discovery, Surveillance and Intelligence, Unmanned Air Systems Projects were found related to the subject.[1]

The image processing system of the researched project, which is written in C++ programming language, is activated automatically without any human assistance. When the UAV geographically enters the search area, the image processing system automatically starts. The image processing algorithm uses the OpenCV library, a powerful and well-known tool for image processing. The images sent to the UAV ground station are processed.

For target detection, the existence of the target object within the acquired image is checked firstly. For this purpose, noise reduction algorithms are applied to the image, converting it to HSV color space format. After the conversion, MSER Blob Detection Algorithm is applied, revealing target candidates considering their aspect ratios.

For the autonomous flight function, Auto-Pilot System Pixhawk was used. Pixhawk is an open source software which allows its' user to make changes to the source via its interface.

To put the idea into practice, the classification and detection of various live objects viewed by using an unmanned aerial vehicle was decided to be performed. In the hardware part of the study, a DJI Phantom 3 Professional UAV was used to obtain images and a Samsung Galaxy S2 tablet PC was used to control the UAV and process the images received.

Various researches have been carried out to resolve the software needs of the study. To work out the goals, it was decided to integrate the classification and detector features of the Tensorflow Android Camera Demo application with the study. For the integration process; Android Studio development environment, Android NDK toolkit, Gradle, CMake and Ndk-Build compilers were used. Android Camera Demo is an open source application shared by the Tensorflow community via Github[14].

Tensorflow is a very powerful and open source deep learning library developed by the Google Brain Team [16]. The library, written in Python and C ++ languages, supports a large variety of operating systems.[13] With Tensorflow, a computer becomes capable of object classification, object detection, voice recognition, cluster analysis and much more.

The classification feature of the Tensorflow Android Camera Demo application sorts the identified objects in the frame according to ascending prediction rate, listing the top 3 together with their precision percentages at the upper part of the screen. The Detector feature detects and marks the identified objects within the image.

Tensorflow libraries written in C++ programming language was compiled during the development of the application together with the FFMPEG library which also depends on C++ language. The FFMPEG library was used during the decoding of the frame obtained from the UAV.

DJI Mobile SDK Library was used for the communication of the developed application with the UAV. The library supports devices with Android or iOS operating systems. The developed application was registered to the DJI Developer System together with the developer information and intended use in order to gain access permit to the SDK.

The rest of the paper is structured as follows; the developed system is demonstrated at Section II and the experimental results are shown at Section III.

## II. DEVELOPED SYSTEM

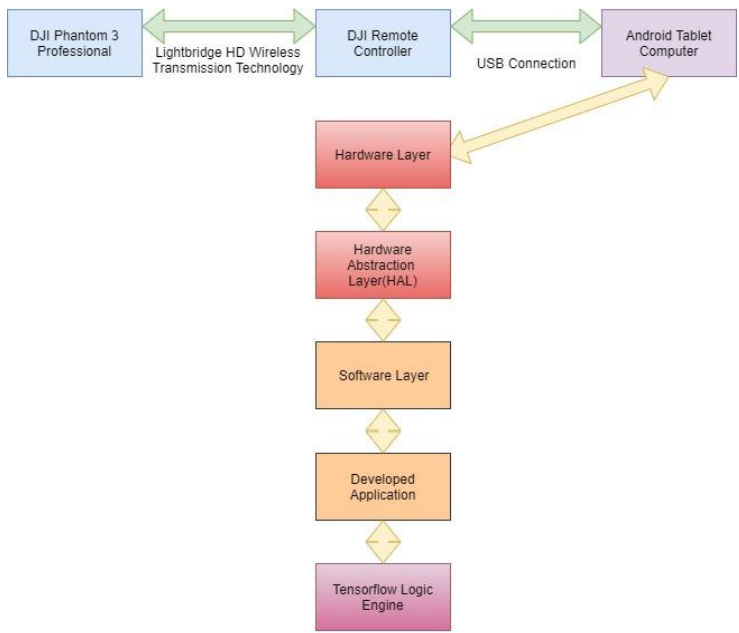

Fig. 1 Developed System's Hierarchical Block Diagram

### A. DJI SDK Library Activation

In order for the developed application to obtain image frames from the UAV, DJI Mobile SDK Library was used. The developed application was activated using the DJI API Key, to gain access to the library. To acquire the API key, a developer account was registered to the DJI Developer system, where the name of the developed application, package name and purpose of use was specified to register the application yielding the 24-digit API activation key.

The resulting key is then included in the software development process as a meta-data through the agency of the AndroidManifest.xml file. Library activation is performed by `ConnectionActivity` class of the 'app' module which is the main component of the study.

### B. Tensorflow Library

Tensorflow library calculations are done using data flow diagrams. In this process, diagram nodes represent mathematical operations and diagram vertices represent data strings (tensors) that represent inter-node flow. The usage areas of the system are quite flexible. Image classification, object detection, sound classification, pattern recognition and pattern generation according to a specific rule are just a few of these areas of use.

Tensorflow library allows machine learning through artificial neural networks. The logic engine is trained and conclusions are made with the experience gained from the training. The input data received during application is directed to the Tensorflow layers. Neurons within these layers may be activated according to the input data. If the input image reaches the top layer by activating the neurons within lower layers, a judgment about the content of the frame can be made.

An image database pre-trained by the Tensorflow community was used in the study. This database contains images categorized according to their contents. The database, which contains at least 500 images in each category, has become a standard of image processing performance. The success of an image processing application is measured by its performance on the imagenet[15] database. There are 14.197.122 images in the database at the moment.

### C. Obtaining Permissions Required from the Android Operating System

The developed application uses features of the Android device such as the internet access, location services, camera, storage read and write, screen timeout prevention and USB port access. Due to the Android operating system security procedures, the application had to be authorized to access these features. To do so, features needed were declared at the Android Manifest file. After that, the device user was prompted to grant required permissions via a popup.

### D. Application Design

The visual interface of the developed application is created through the activity classes it contains. Each activity class retrieves the design of the user interface it uses from an XML file. This file defines the interface elements and their locations. The interface elements are manipulated via Java code found within the relevant activity class.

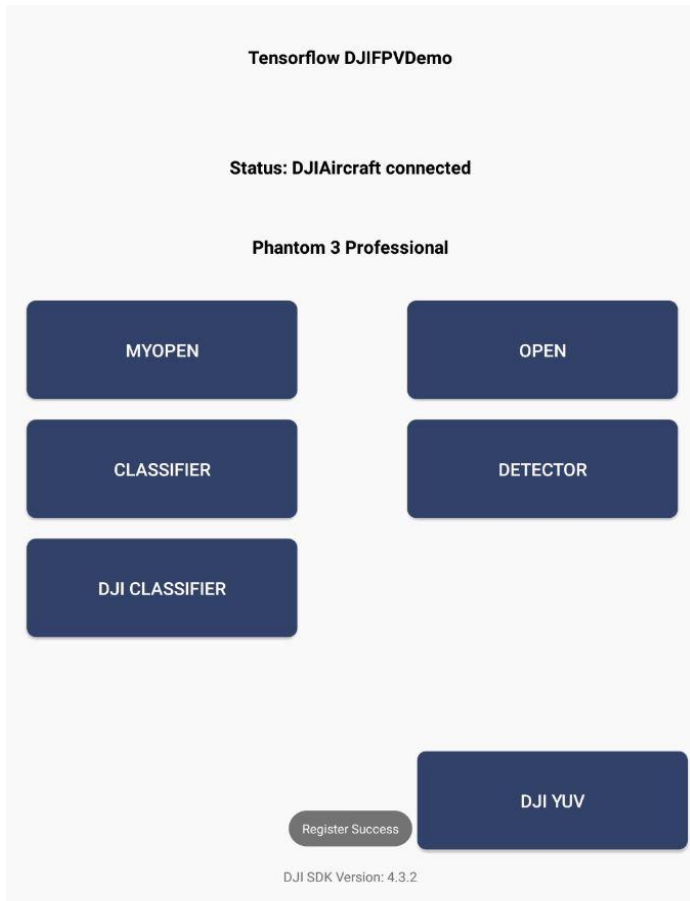

Fig. 2 Connection Activity

The developed application's home page, as shown Figure 2, is implemented by `ConnectionActivity` class. This class fulfills the DJI SDK library activation, connection to the UAV control unit and the retrieval of the UAV information. The result of the library activation process is notified to the end user via a Toast message. The UAV connection information and the version of the connected UAV are displayed on the activity through textview elements. In addition to these functions, another feature of the home screen is that it provides access to other activities via buttons. Desired activity is triggered after the user presses one of the buttons.

In the classification activity, image frames taken from the UAV is processed to classify the objects within the frame. The largest element of this activity is the textview element as can be seen in Figure 3. Through this element, images taken from the UAV is displayed to the user. The result of the image processing operation is presented to the user through the blue `RecognitionScoreView` view element located at the top of the activity. The first three predictions with the highest accuracy are listed in the `RecognitionScoreView` element.

The image, which is displayed via the textview object, is only used for enhancing the user experience. Image processing operation is done in the background, by decoding received image frames. The classification process takes at least 300 milliseconds. In the meantime, the user is presented with a dynamic screen via the `textview` element.

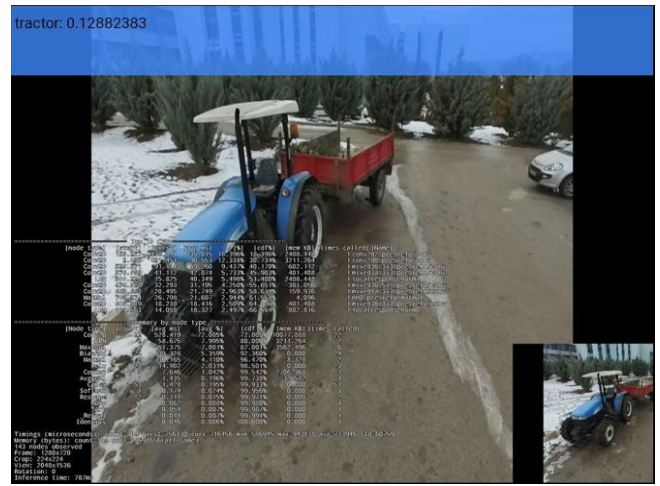

Fig. 3 Classifier Activity with Details Layer Activated

The details are viewed by adding a layer with invisible background over the existing activity. It is activated and deactivated by pressing the 'volume down' key of the Android device. At the lower right corner of the screen, decoded image which was sent to the Tensorflow Image Classification Function is displayed. At the lower left corner, resolution of the image frame taken from the UAV, resolution of the image sent to the Tensorflow function, resolution of the device's screen, camera's rotation degree and the time elapsed during the processing of the last image frame are displayed. In the center of the screen, the top 10 Tensorflow functions, which occupy most of the processor time, are shown in table format with information of all other functions used.

Classification activity inherits `CameraActivity` class. In this class, YUV image data which is broadcasted by DJI image stream decoder is captured through `IYuvDataListener`, depicted in Figure 4. After relevant format conversions, the image is sent to Tensorflow logic engine.

```
DJIVideoStreamDecoder.getInstance().setYuvDataListener(CameraActivity.this);
```

Fig. 4 Registering the CameraActivity class as YUV Data Listener

```java
// The callback for receiving the raw H264 video data for camera live view
mReceivedVideoDataCallBack = new VideoFeeder.VideoDataCallback() {

  @Override
  public void onReceive(byte[] videoBuffer, int size) {/*
!!!!!!!!!!!!!!!! FRAME RECEIVED !!!!!!!!!!!!!!!!!!!!!!!!*/
 //Log.i(TAG, "VideoDataCallBack videoBuffer Received of H.264 TYPE");
    if (mCodecManager != null) {
        DJIVideoStreamDecoder.getInstance().parse(videoBuffer, size);//sends video to decoder for YUV(MUST BE BEFORE SURFACE DECODER!!!)
        mCodecManager.sendDataToDecoder(videoBuffer, size);//flushes videobuffer !!MUST BE AFTER YUV DECODER PARSER!!
    }
  }
};
```

Fig. 5 Sending Image acquired from the UAV Camera to Decoders

Detector Activity also inherits `CameraActivity` class like the Classification activity. The difference is that instead of the Tensorflow Classify function, the Tensorflow Yolo Detector function is called to identify the objects previously taught in the database by training and to mark them in the `textview` element.

*E. Image Acquisition*

Within the `CameraActivity` class, an object of the `LegacyCameraConnectionFragment` class is created. The camera functions are carried out through this object. The raw H264 formatted image is captured and sent to both `DJIVideoStreamDecoder` and `DJICodecManager` [12] decoder. `DJIVideoStreamDecoder` broadcasts image in YUV data format and `DJICodecManager` feeds the `textview` element used to present the image to the user.

*F. Image Processing*

The `CameraActivity` class captures the YUV data broadcasted by the `DJIVideoStreamDecoder` via the `onYuvDataReceived` function of the `IYuvDataListener` event listener. First of all, the function checks whether a previously started image processing process is continuing. If the processing of the previous image frame is not completed, the function is terminated via the return command and the current YUV data is dropped as shown in Figure 6.

```
@Override
public void onYuvDataReceived(final byte[] yuvFrame, int width,
int height) {
/*  Log.d(TAG,
        "onYuvDataReceived: frame index: "
                +
DJIVideoStreamDecoder.getInstance().frameIndex);*/

  if (isProcessingFrame) {
    LOGGER.w("Dropping frame!");
    return;
  }

  final byte[] bytes = YuvToYuv420Sp(yuvFrame, width, height);
```

Fig. 6 State Control of Image Processing and Triggering of Format Transformation

As YUV data are received, it is checked whether an image processing event is occurring or not. If not, the image obtained in YUV data format via the listener is converted to ARGB8888 image format. After that, image is sent to the Tensorflow logic engine. The function finally indicates that there is data ready to be processed to the Tensorflow logic engine, by triggering the `processImage` function. Doing so, the processing of the image frame begins.

### III. EXPERIMENTAL RESULTS

After the application was developed, a test run was made at Ankara University Computer Engineering Department's parking lot. Objects such as tractor, lake, pier and a variety of vehicle types (racer, sports car, convertible and etc.) were classified by the classifier activity. Vehicles were also detected and marked by the detector activity of the developed application. For testing purposes, a few cat pictures, toy taxi, keyboard laptop computer, UAV controller's joystick were displayed to the UAV's camera manually and all of the so mentioned were spotted by the developed application.

TABLE I. CLASSIFICATION ACCURACY TEST

| Object | Accuracy |
|---|---|
| Joystick | 19.04% |
| Laptop | 11.68% |
| Laptop Monitor | 14.76% |
| Tractor | 12.88% |
| Sports Car | 28.74% |
| Racer | 27.16% |
| Convertible | 13.45% |
| Cars in General | 23.17% |
| Cat | 22.79% |
| Seashore | 24.44% |

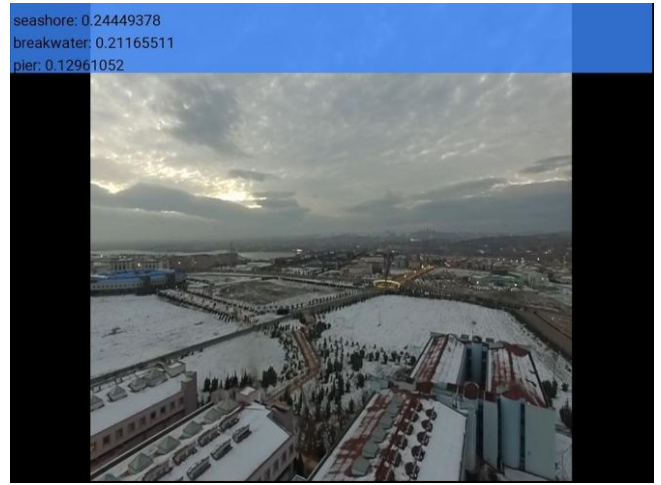

Fig. 7 Classification Activity Sea Edge Detection
(UAV Hovering Over Ankara University Computer Engineering Department)

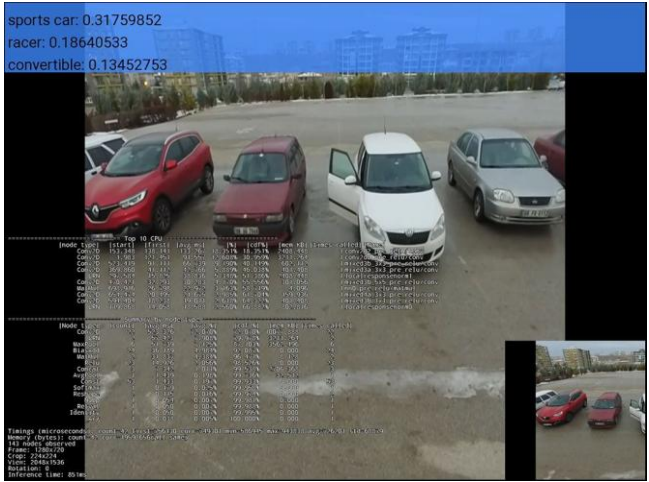

Fig. 8 Classification Activity Vehicle Detection
(UAV Hovering Over Ankara University Computer Engineering Department Parking Lot)

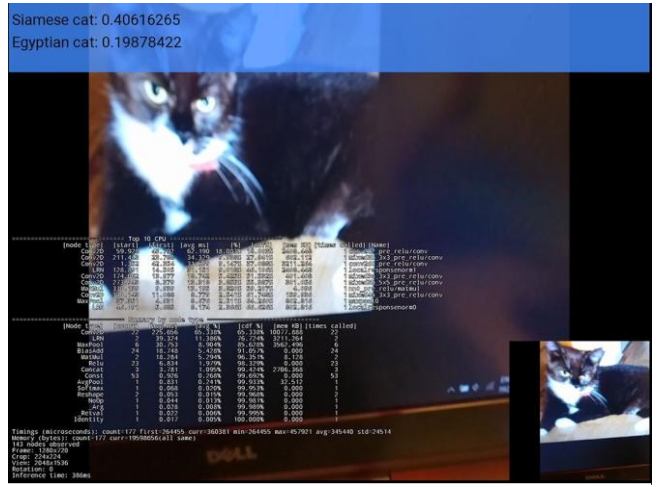

Fig. 11 Classification Activity Cat Detection

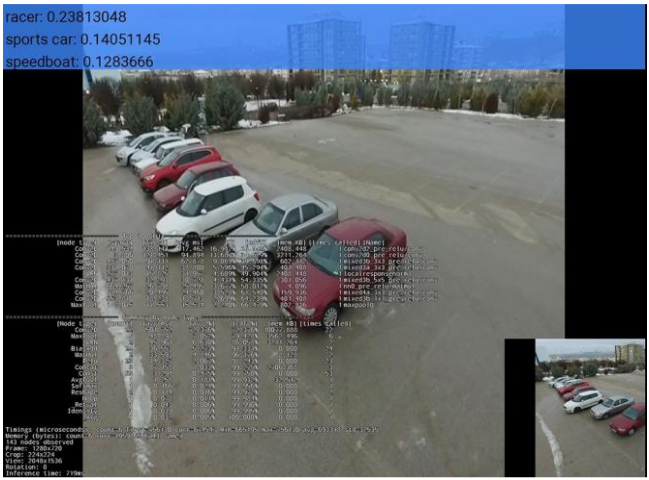

Fig. 9 Classification Activity Vehicle Detection with Details Layer Activated
(UAV Hovering Over Ankara University Computer Engineering Department Parking Lot)

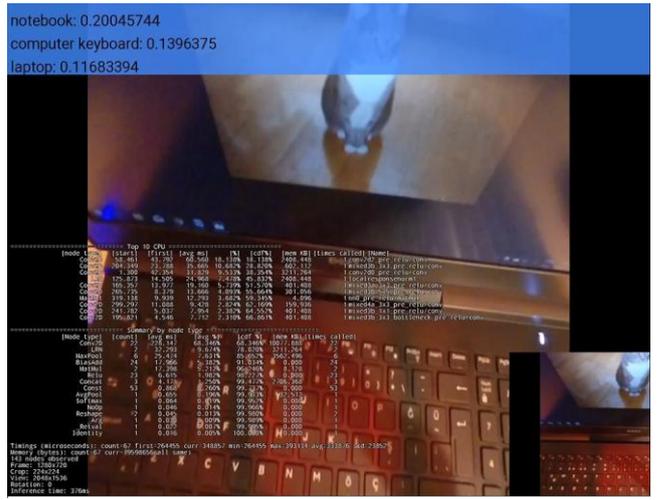

Fig. 12 Classification Activity Notebook Computer Detection

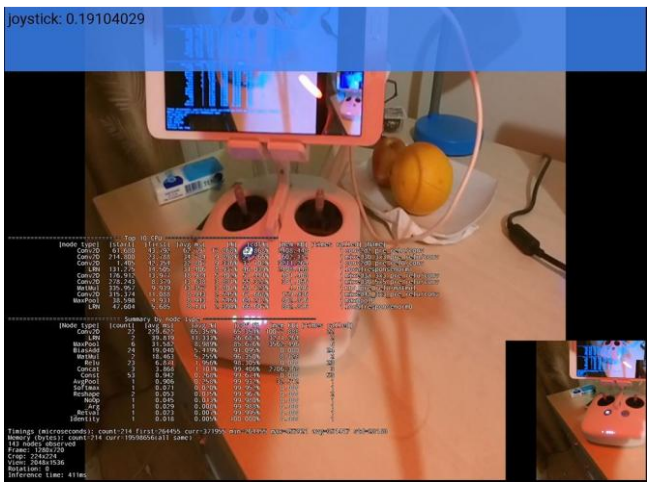

Fig. 10 Classification Activity Controller Unit's Joystick Detection

IV. CONCLUSION

The intention of this study was to develop an Android application, which analyzes and interprets objects within the image frames, acquired by DJI Phantom 3 Professional UAV. Various researches have been carried out on the subject and it was decided to use Artificial Neural Networks trained by Machine Learning for the analysis and interpretation of the images. To do so, the Tensorflow sample Android application which includes all the necessary functions and libraries was used during application development.

Integration of Tensorflow sample Android application with DJI Phantom 3 Professional UAV was made in the scope of this study. The pre-trained Inception-V3 model was used for classification and Mobilenet model was used for detection of objects. As a result, an Android application that acquires image frames from the UAV, classifying objects up to 1000 categories and detecting up to 171 different types of objects was discovered.

The developed application is planned to be improved during future studies. First of all, it is aimed to reconfigure the process of transmitting the acquired image to the Tensorflow logic engine, reducing the number of image format conversions. After that, the calibration problem occurring during marking of the detection objects will be solved and the DJI application widgets will be added to the application. Finally after these objectives are satisfied, certain objects will be marked with an attached laser pointer after being detected by the system.

REFERENCES

[1] Öznalbant Z. Filik T. Gerek Ö.N. "KGİ(Keşif, Gözetleme ve İstihbarat) İHS(İnsansız Hava Sistemleri) Projeleri Kapsamında Anadolu Üniversitesinde Yapılan Çalışmalar". IX. TURKISH NATIONAL AIRCRAFT AEROSPACE ENGINEERING CONGRESS PAPERS, 168-171

[2] Abadi, Barham, Chen, Proceedings of the 12th USENIX Symposium on Operating Systems Design and Implementation (OSDI '16).2016. TensorFlow: A System for Large-Scale Machine Learning.; 265-278

[3] Goldsborough P., A Tour of TensorFlow, 2016.

[4] Abeywickrama S., Jayasinghe L., Fu H., Yuen C., ICASSP 2018. RF-based Direction Finding of Small Unmanned Aerial Vehicles Using Deep Neural Networks

[5] Baylor D., Breck E., Cheng H., Fiedel N., Foo C. Y., Haque Z., Haykal S., Ispir M., Jain V., Koc L., Koo C. Y., Lew L., Mewald C., Modi A. N., Polyzotis N., Ramesh S., Roy S., Whang S. E., Wicke M., Wilkiewicz J., Zhang X., Zinkevich M., GOOGLE INC., KDD 2017 Applied Data Science, TFX: A TensorFlow-Based Production-Scale Machine Learning Platform

[6] Kovalev V., Kalinovsky A., Kovalev S., Minsk: Publishing Center of BSU PATTERN RECOGNITION AND INFORMATION PROCESSING (PRIP'2016). Deep Learning with Theano, Torch, Caffe, Tensorflow, and Deeplearning4J: Which One is the Best in Speed and Accuracy?

[7] Howard A. G., Zhu M., Chen B., Kalenichenko D., Wang W., Weyand T., Andreetto M., Adam H., GOOGLE INC. 2017. MobileNets: Efficient Convolutional Neural Networks for Mobile Vision Applications

[8] Ivanova D., Kadurin V., Proceedings of the International Conference on Information Technologies (InfoTech-2017). Mobile Application For Pattern Recognition Based On Machine Learning

[9] Sheppard, Rahnemoonfar, School of Engineering and Computing Sciences, Texas A&M University-Corpus Christi (Access Year: 2017). Real-time Scene Understanding for UAV Imagery based on Deep Convolutional Neural Networks

[10] Yoo J., Hong Y., Yoon S. (Access Year: 2017). Autonomous UAV Navigation with Domain Adaptation

[11] "CS 20SI: Tensorflow for Deep Learning Research", access time 30 December 2017, https://web.stanford.edu/class/cs20si/

[12] "DJICodecManager", access time 30 December 2017, https://developer.dji.com/iframe/mobile-sdk-doc/android/reference/dji/sdk/Codec/DJICodecManager.html

[13] "Top Five Use Cases Of Tensorflow", access time 30 December 2017, https://www.exastax.com/deep-learning/top-five-use-cases-of-tensorflow/

[14] "TensorFlow", access time 30 December 2017, https://github.com/tensorflow/tensorflow

[15] "ImageNet ", access time 30 December 2017, http://image-net.org/index

[16] "Google Brain Team", ", access time 21 December 2017, https://research.google.com/teams/brain/